\documentclass[conference]{IEEEtran}
\IEEEoverridecommandlockouts
\usepackage{cite}
\pdfoutput=1
\usepackage{amsmath,amssymb,amsfonts}
\usepackage{graphicx}
\usepackage{textcomp}
\usepackage{xcolor}
\usepackage{tikz}
\usepackage{overpic}
\usepackage{amsmath}

\DeclareMathOperator*{\argmin}{arg\,min}

\usepackage[linesnumbered,ruled,vlined]{algorithm2e}
\usetikzlibrary{automata,arrows,positioning,calc}
\def\BibTeX{{\rm B\kern-.05em{\sc i\kern-.025em b}\kern-.08em
    T\kern-.1667em\lower.7ex\hbox{E}\kern-.125emX}}

\SetCommentSty{mycommfont}

\SetKwInput{KwInput}{Input}                
\SetKwInput{KwOutput}{Output}              
\begin{document}

\title{Unreliable Multi-Armed Bandits: A Novel Approach to Recommendation Systems\\
\thanks{*Equal contribution, first author was assigned randomly}
}

\author{\IEEEauthorblockN{Aditya Narayan Ravi*}
\IEEEauthorblockA{
\textit{IIT Bombay}\\
Mumbai, India \\
aditya\_narayan@iitb.ac.in}
\and
\IEEEauthorblockN{Pranav Poduval*}
\IEEEauthorblockA{ 
\textit{IIT Bombay}\\
Mumbai, India \\
150110007@iitb.ac.in
}
\and
\IEEEauthorblockN{Sharayu Moharir}
\IEEEauthorblockA{ 
\textit{IIT Bombay}\\
Mumbai, India \\
sharayum@ee.iitb.ac.in}
}

\maketitle

\begin{abstract}
We use a novel modification of Multi-Armed Bandits to create a new model for recommendation systems. We model the recommendation system as a bandit seeking to maximize reward by pulling on arms with unknown rewards. The catch however is that this bandit can only access these arms through an unreliable intermediate that has some level of autonomy while choosing its arms. For example, in a streaming website the user has a lot of autonomy while choosing content they want to watch. The streaming sites can use targeted advertising as a means to bias opinions of these users. Here the streaming site is the bandit aiming to maximize reward and the user is the unreliable intermediate. We model the intermediate as accessing states via a Markov chain. The bandit is allowed to perturb this Markov chain. We prove fundamental theorems for this setting after which we show a close-to-optimal Explore-Commit algorithm.
\end{abstract}





\section{Introduction and Problem Formulation}
Multi-Armed Bandits and algorithms to characterize the exploration-exploitation trade-off have been studied for a long time in various literature \cite{Auer2002,doi:10.1002/asmb.874}. It's use in recommendation systems has also been studied in various literature \cite{6483433,10.1007/978-3-642-34487-9_40}. \\
However most literature analyses recommendation systems from a data-acquisition and use perspective, usually fails to address the question on how whether the cost incurred in collecting and using this data for recommendation systems is worth the reward generated from how they bias autonomous user choices. We attempt to do that by developing the \textbf{Unreliable Multi-Armed Bandits} framework, and prove a few fundamental bottlenecks on the cost incurred by recommendation systems\\
A $K$-armed bandit is defined by Random Variables (RV) $ \mathbf{R_{i,j}}: 1 \leq i \leq K, 1 \leq j $. The RVs $\mathbf{R_{i,j}}$ refers to reward produced by arm $i$ at the $j^{th}$ time instant. Here $\mathbf{R_{i,j}}$ is independent of $\mathbf{R_{s,t}}$ if $i \neq s, j \neq t$ and additionally identically distributed if $i = s, j \neq t$. These rewards are governed by an unknown underlying distribution (hence with an unknown mean $\mu_{i}$). The optimal arm is the arm such that $\text{max} \, \mu_{i} = \mu^{*}$ is mean reward of that arm.\\
Let an unreliable intermediate $I$ visit different arms by traversing an irreducible and aperiodic Markov chain with state space $S$ and a transition probability matrix $P = [p_{i,j}]_{|S| \times |S|}$.
\begin{figure}
\begin{center}
\begin{tikzpicture}[->, >=stealth', auto, semithick, node distance=3cm]
\tikzstyle{every state}=[fill=white,draw=black,thick,text=black,scale=1]
\node[state]    (1)                     {$1$};
\node[state]    (2)[right of=1]         {$2$};
\path
(1) edge[loop left]    node{$p_{1,1}$}         (1)
(1) edge[bend left]    node{$p_{1,2}$}         (2)
(2) edge[bend left]    node{$p_{2,1}$}         (1)
(2) edge[loop right]    node{$p_{2,2}$}         (2);
\end{tikzpicture}
\end{center}
\begin{center}
\begin{tikzpicture}[->, >=stealth', auto, semithick, node distance=3cm]
\tikzstyle{every state}=[fill=white,draw=black,thick,text=black,scale=1]
\node[state]    (1)                     {$1$};
\node[state]    (2)[right of=1]         {$2$};
\path
(1) edge[loop left]    node{$p_{1,1} - \delta$}         (1)
(1) edge[bend left]    node{$p_{1,2} + \delta$}         (2)
(2) edge[bend left]    node{$p_{2,1} - \delta$}         (1)
(2) edge[loop right]    node{$p_{2,2} + \delta$}         (2);
\end{tikzpicture}
\end{center}
    \caption{2-state Markov chain a) Without the effect of recommendation system b) With the effect of recommendation system}
    \label{figure:Markov}
\end{figure}
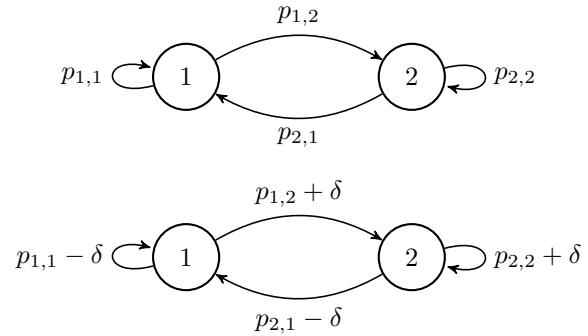
 The state space S is the set of arms and has a cardinality $|S| = K$ and $P$ models the autonomy of the intermediate in pulling the arms. This Markovian structure shows the dependence of future choices of the intermediate based on the current choice. For example a person who watches a “Horror” movie and likes it, is likely to pick the next movie they watch as a “Horror” movie as well. In our analysis we make the assumption that $P$ is known. This is justified because recommendation systems usually possess enough data about the user to make predictions on the user’s behaviour. Thus even if $P$ is initially unknown, it is easily estimated. Figure \ref{figure:Markov}a) shows the structure for a simplified 2-state Markov chain\\
Let $\delta$ be known as the perturbation variable. This variable is used to model the effect of recommendation systems. The streaming site can use methods like ``targeted advertising" to bias the user choices (i.e. Suggesting they pick another movie, say a ``Documentary" which increases the chance they pick that movie). Therefore $\delta$ is used to perturb the probability of transition from a particular state. The following rules regarding $\delta$ is followed,
\begin{itemize}
    \item This $\delta$ is added and the transition probabilities are suitably modified to maintain a sum to one
    \item If this modification causes negative probabilities (probabilities greater than $1$) to arise, the probabilities are suitably truncated to 0 ($1$)
    \item We make the simplifying assumption that $\delta$ is a pre-fixed parameter, future work will deal with more realistic scenarios

\end{itemize}
Figure \ref{figure:Markov}b) shows the perturbed version of the Markov chain\\
Let $\pi \in \mathbf{\Pi}$, be a perturbation policy which specifies the perturbation of $P$. We define Regret as $\Delta_{j} = \mu^{*} - r_{j}$, where $r_{j}$ refers to the real reward obtained at time instant $i$. Therefore the Cumulative Regret ($C_{T}$) after $T$ plays / biases is,
\begin{align}
   C_{T} = \sum_{j = 1}^{T} \Delta_{i} = T\mu^{*} - \sum_{j = 1}^{T}r_{j}.
\end{align}

Here $\pi^{*}$ is the optimal policy, (i.e. the one that has least cumulative regret in expectation). The goal is to minimize the $C_{T}$ over $T$.\\
The structure of the paper is as follows: First we state 2 theorems on the regret bound and optimal policy of our setup. Following the proofs of these theorems, we look at a heuristic explore-commit algorithm based inspired by these theorems and show (via simulations) that it is near optimal for enough number of states (cardinality of state space is reasonably large)
\section{Main Results}
Our first theorem characterizes which policy is optimal to maximize reward obtained. We first define a few concepts to make the theorem clear.\\
\textit{Definition 1.} The bandit is said to have a ``genie" available if $\mu_{i}$ is known to the bandit for all $i \in [1:K]$ \\
\\
\textit{Definition 2.} (When genie is available) A $\pi^{\delta^{*}}$ policy is a policy that perturbs $P$ to $P^{\delta^{*}}$, where $P^{\delta^{*}}$ is the modified transition probability matrix that maximizes the probability of transitioning to the state with maximum mean reward (from any state)\\
\\
\textbf{Theorem 1.} The optimal policy (when genie is available) $\pi^{*} \in \mathbf{\Pi}$ is $\pi^{\delta^{*}}$  \\
\\
The proof for this theorem for the case where the recommendation system happens after the perturbed Markov chain attains it's stationary distribution. This is easily justified since we can allow the Markov chain to evolve from $T = -\infty$ and start our observations from $T = 0$ The proof generalises to the finite case (where the Markov chain doesn't attain it's stationary distribution) quite easily, but we exclude it for brevity. The importance of this theorem is that it provides us a heuristic to construct the explore-commit algorithm discussed in Section IV. \\
Our next theorem shows that there exists a fundamental bottleneck on the rewards that can be extracted from the setup. This is intuitively due to the fact that even if the bandit knows exactly which state gives what reward, the bandit doesn't have full control over which arm it can pull at each instant. The following theorem asserts that this bottleneck is in fact linear. \\
\\
\textbf{Theorem 2.} For any $\pi \in \mathbf{\Pi}$, the $C_{T}$ can be lower bounded (in expectation) as,
\begin{align}
    \mathbf{E}\left(C_{T}\right) \geq T\left( \mu^{*} - \Tilde{\mu}\right).
\end{align}
where $\Tilde{\mu}$ is the sum of the means of each the underlying distribution of each arm, weighted by the stationary probability distribution of the transition probability matrix after $\pi^{\delta^{*}}$ policy is applied on the Markov Chain \\
Again for brevity the proof technique in the next section assumes that the perturbed Markov chain attains it's stationary distribution. The theorem holds true in order (both are lower bounded by functions which are linear in $T$) even otherwise as well, though some of the constants vary. The importance of this theorem is that it gives us a metric to evaluate our algorithm against. The explore-commit algorithm described in Section IV is observed to be near optimal and robust with respect to this theorem.

\section{Proofs}
Let $\boldsymbol{\nu} = [\nu_{i}],i \in [1:K]$ be a vector representing the stationary probability distribution of the Markov chain. This is defined as the solution to the equation $\nu P = \nu$. This equation has a unique solution since the Markov chain is irreducible and aperiodic. When T is infinity this stationary distribution defines the amount of time (in expectation) spent in each state\\
\\
\textbf{Proof of Theorem 1:}
Without loss of generality assume $\mu_{i}$ is a decreasing function of $i$\\
We prove this by induction on the number of states $K$ of the Markov chain \\
\textit{Base case:} Let $K = 2$. Then the total reward ($R$) in $T$ steps is,
\begin{align}
    R = \sum_{i=1}^{T} r_{i}
\end{align}
\begin{align} \label{2-case}
    \mathbf{E}(R) = \sum_{i = 1}^{T}(\nu_{1}\mu_{1} + \nu_{2}\mu_{2}) = T(\nu_{1}\mu_{1} + \nu_{2}\mu_{2}).
\end{align}
Since $\nu_{1} + \nu_{2} = 1$, We can rewrite (\ref{2-case}) as \\
\begin{align}
     \mathbf{E}(R) =  T(\nu_{1}\mu_{1} + (1 - \nu_{1})\mu_{2}) = T(\nu_{1}(\mu_{1} - \mu_{2}) + \mu_{2}).
\end{align}
Since $\mu_{1} \geq \mu_{2}$, maximizing $\nu_{1}$ is our goal \\
It is easy to see that policy $\pi^{\delta^{*}}$ maximizes $\nu_{1}$ (and hence the expression). \\
\textit{Inductive step:} Assume the theorem is true for $K = l - 1$ states. We now prove that the theorem is true for $K = l$ states. Note that,
\begin{align} \label{l-case}
    \mathbf{E}(R) = T\left(\sum_{j=1}^{l-1}\nu_{j}\mu_{j} + \nu_{l}\mu_{l}\right).
\end{align}
Now (\ref{l-case}) can be rewritten as,
\begin{align} \label{l-case-rewritten}
    \mathbf{E}(R) = T\left(\sum_{j=1}^{l-1}\nu_{j}(\mu_{j} - \mu_{l}) + \mu_{l}\right).
\end{align}
Take the first term in (\ref{l-case-rewritten}). We can normalise this by dividing this term by $B =   \sum_{j=1}^{l-1}\nu_{j} = 1 - \nu_{l}$. Let $\frac{\nu_{i}}{B} = \nu_{i}^{*}$. There $\exists$ a Markov chain with $l - 1$ states that has a unique stationary probability distribution $\boldsymbol{\nu^{*}}$.\\
By induction hypothesis, this new Markov Chain is maximised by a $\pi^{\delta^{*}}$ policy. Note here that B doesn't depend $\nu_{i}, i \in [1:l-1]$ when $\nu_{l}$ is specified. Since (\ref{l-case-rewritten})
only depends on $\nu_{i}, i \in [1:l-1]$, the $\pi^{\delta^{*}}$ policy applied on $\boldsymbol{\nu^{*}}$ also maximises (\ref{l-case-rewritten}). This proves the first theorem. A similar analysis also follows for a finite $T$ (though the calculations are a bit cumbersome)\\
\\
\textbf{Proof of Theorem 2:} We note the following facts
\begin{itemize}
    \item If a genie is available, the cumulative regret of the original problem can only go down, in expectation
    \item When genie is available, Theorem 1 states that a $\pi^{\delta^{*}}$ policy is optimal
    \item A $\pi^{\delta^{*}}$ policy perturbs $\boldsymbol{\nu}$ to $\boldsymbol{\nu^{\delta}}$
    \end{itemize}
The above observation tells us that calculating $\mathbf{E}(C_{T})$ when genie is available, for the $\pi^{\delta^{*}}$ policy will give us a lower bound on $\mathbf{E}(C_{T})$ for any general policy and unknown reward distributions. Let the setting required be $\mathbf{E}(C_{T}^{*})$.\\
To this end lets calculate $\mathbf{E}(C_{T}^{*})$.
\begin{align}
    \mathbf{E}(C_{T}^{*}) & =\sum_{i = 1}^{T}\left( \sum_{j = 1}^{K}\nu_{j}^{\delta}(\mu^{*} - \mu_{j})\right) \\
    & = T\left(\mu^{*} - \sum_{j = 1}^{K}\nu_{j}^{\delta}\mu_{j}\right)\\
    & = T(\mu^{*} - \Tilde{\mu}).
\end{align}
It is easy to see that $\Tilde{\mu} \leq \mu^{*}$. This proves theorem 2.
\section{Algorithms}
\textit{Definition 3:} A $\pi^{l}$ policy is a policy that perturbs $P$ to $P^{l}$, where $P^{l}$ is the modified transition probability matrix when Markov chain is biased towards state $l \in [1:K]$ \\
We construct the Perturbation Explore ($\text{P}^{2}\text{EE}$), Perturbation Commit algorithm based on insights from Theorem 1. The intuition behind the algorithm is as follows,
\begin{itemize}
    \item We explore ``enough" so that we get a distinguishable estimate of the means of each of the arms. To do this we perturb the Markov chain towards the arm that is least explored at a given time instant (i.e. most uncertain)
    \item  Since after exploration we are pretty much the ``genie" now, theorem 1 kicks in. We commit to the arm with highest empirical reward by using the $\pi^{\delta^{*}}$ policy
\end{itemize}
\begin{algorithm}
\SetAlgoLined
\textbf{Inputs: }{ K, T, \textit{Transition Probability Matrix}}\\
\textbf{Initialize: }{\textit{Pulls} = $[0, \dots, 0 ]$, $t = 0$}\\
 \While{$t \leq KT$}{
  \textit{\textbf{Exploration:}}
\textit{bias} = $\argmin$ \textit{Pulls} \\
$\pi^{bias} \in \mathbf{\Pi}$

 }
 \While{$KT \leq t  \leq T$}{
 \textbf{Exploitation:}
$\pi^{\delta^{*}} \in \mathbf{\Pi}$
 }
 \caption{$\text{P}^{2}\text{EE}$}
\end{algorithm}
\textbf{Note:} We haven't characterized exactly how to choose $K$ in the above algorithm. Here we just choose $K$ manually based on simulations. Future work will deal with proving a theoretically optimal choice of $K$. \\
In the upcoming section we compare our algorithm two other algorithms, Upper Confidence Bound (UCB) \cite{Auer2002} and a Greedy algorithm. But slight modifications need to be made to these algorithms to fit our problem statement. \\
The UCB algorithm remains same, except that in our case we cannot guarantee that we pick the arm with the most confidence in each time-step. Rather we can only bias the Markov chain so that it is more likely to pick said arm. Therefore we compare our algorithm keeping this in mind. \\
The vanilla $\epsilon$-Greedy algorithm \cite{Auer2002} chooses a random arm with probability $\epsilon$. The structure of our Markov chain allows this exploration anyway, since there is always a non-zero chance to pull some sub-optimal arm.
\begin{figure*}[t]
    \begin{minipage}[t]{0.3\textwidth}
    \centering
    \includegraphics[height=1.8in]{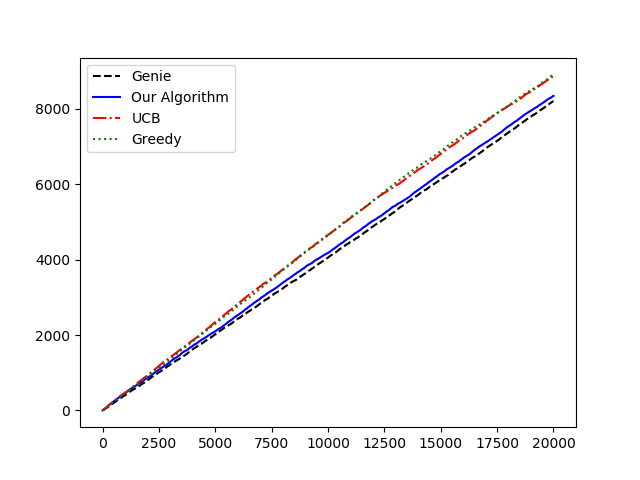}
    \begin{tikzpicture}
        \put (-85,85) {\small{$C_{T}$}}
        \put (0,10) {\small{$T$}}
        \put (-10,0) {\small{$\delta = 0.1$}}
    \end{tikzpicture}
    \end{minipage}%
    ~ 
    \begin{minipage}[t]{0.3\textwidth}
        \centering
        \includegraphics[height=1.8in]{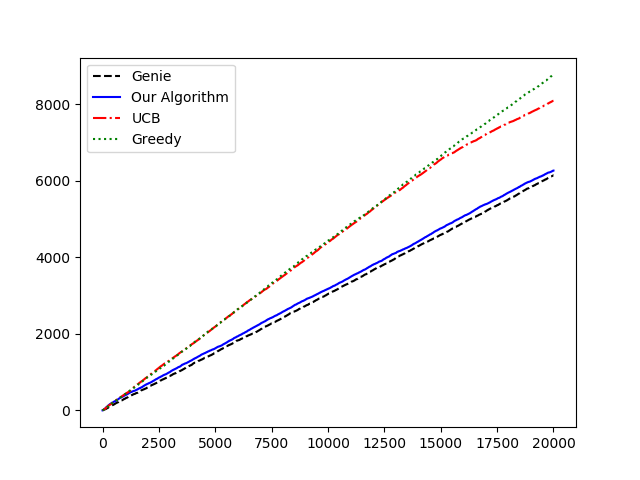}
        \begin{tikzpicture}
            \put (0,10) {\small{$T$}}
            \put (-10,0) {\small{$\delta = 0.3$}}
        \end{tikzpicture}
    \end{minipage}
    ~
    \begin{minipage}[t]{0.3\textwidth}
        \centering
        \includegraphics[height=1.8in]{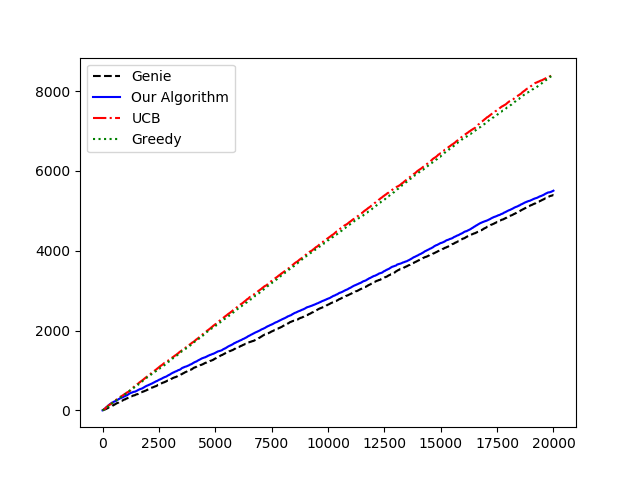}
        \begin{tikzpicture}
            \put (0,10) {\small{$T$}}
            \put (-10,0) {\small{$\delta = 0.5$}}
        \end{tikzpicture}
    \end{minipage}
    \caption{$C_{T}$ vs $T$ comparison for $\text{P}^{2}\text{EE}$,  Our method $P^2$EE has cumulative regret very close to the optimal ``genie" and it's regret decreases as $\delta$ increases. UCB and greedy are clearly sub-optimal }
    \label{figure:Sim}
\end{figure*}
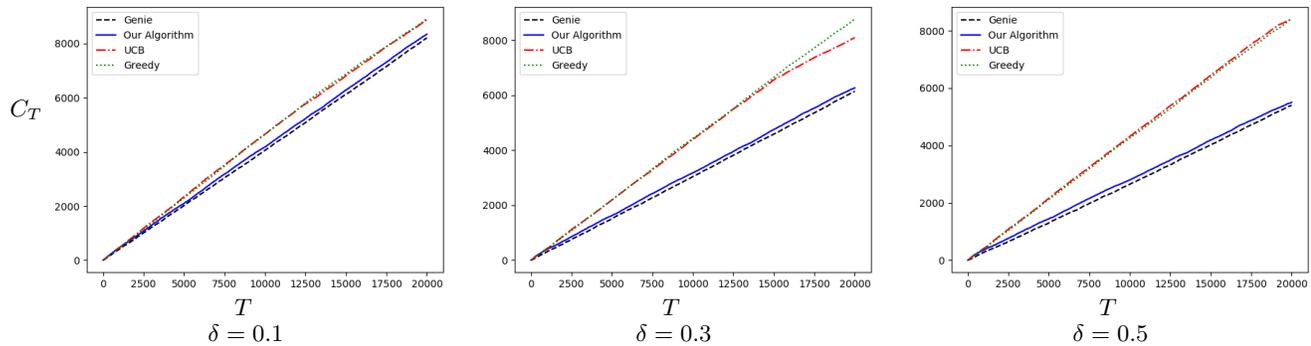
\section{Simulations}
\begin{figure}[htb]
    \centering
    \includegraphics[height=2.7in]{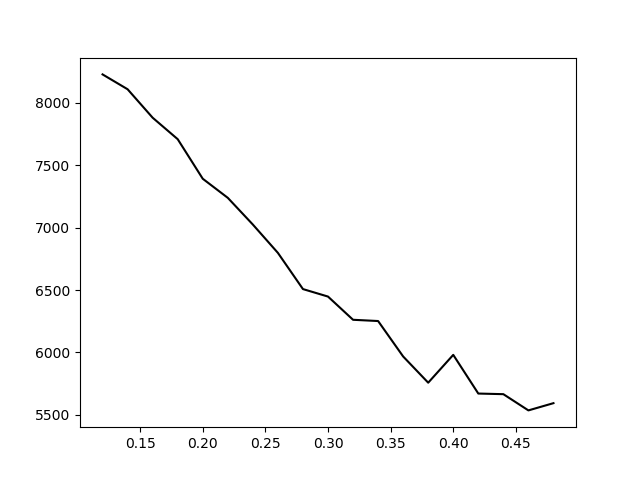}
    \begin{tikzpicture}
            \put (0,10) {\small{$\delta$}}
            \put (-130,100) {\small{$C_{T}$}}
        \end{tikzpicture}
    \caption{Variation of $C_{T}$ as $\delta$ increases for $\text{P}^{2}\text{EE}$}
    \label{fig:Delta}
\end{figure}
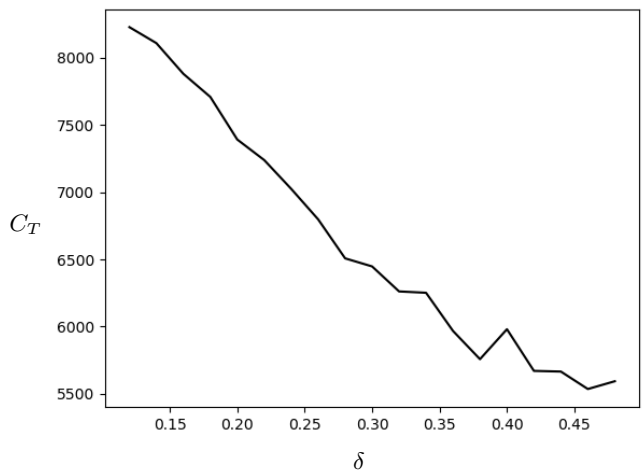

The simulations in Figure \ref{figure:Sim} are done for a 10-state Markov chain with randomly generated transition probability matrix and rewards such that $\mu_{i}$ is a decreasing function of $i \in [1:10]$. As required by the problem statement, the algorithms don't have an access to the rewards Apriori.  \\
We compare our $\text{P}^{2}\text{EE}$ algorithm via simulation with the modified UCB and greedy algorithm displayed in the previous section.  We plot the cumulative regret plots for three different values of $\delta $.\\
Notice that a purely greedy strategy is unreliable as compared to $\mathbf{\text{P}^2\text{EE}}$ method as we have no guarantee the exploration is sufficient to distinguish the optimal arm from the sub-optimal ones. Which could lead us to be pulling the sub-optimal arms for a long time.\\
It is apparent that UCB is a sub-optimal strategy for this setting since it is in violation of our Theorem 1 and is often stuck trying to pull sub-optimal arms for too long. \\$\text{P}^2\text{EE}$ is theoretically supported by Theorem 1 and is empirically very close to the Oracle ``Genie") making it close to optimal.\\
An interesting observation that can be made here is that as the perturbation value $\delta$ increases the cumulative regret decreases for our algorithm. This is consistent with the hypothesis that increasing the influence of the recommendation system will give improved rewards. This is plotted in Figure \ref{fig:Delta}.\\

\section{Conclusion}
Our work has propensity for future work. The first of this would be to generalise the notion of biasing, such that the bias is a) state dependent and b) apply to multiple transition probabilities. The second would be to make the model more realistic in terms of the intermediate's behavior. For example, it is unrealistic to assume that in the example we use that user would watch the same movies multiple times and derive the same reward from it. We therefore need to bring in a rotting bandit \cite{NIPS2017_6900} or restless bandit \cite{6200864} assumption on the rewards. We also haven't taken into account dynamic updations of the Markov chain (because the way users make choices evolves over time). Moreover, there is propensity to extend the probabilistic behavior of the intermediate to more than a Markovian behavior.

\bibliographystyle{plain}

\begin{thebibliography}{1}

\bibitem{Auer2002}
Peter Auer, Nicol{\`o} Cesa-Bianchi, and Paul Fischer.
\newblock Finite-time analysis of the multiarmed bandit problem.
\newblock {\em Machine Learning}, 47(2):235--256, May 2002.

\bibitem{10.1007/978-3-642-34487-9_40}
Djallel Bouneffouf, Amel Bouzeghoub, and Alda~Lopes Gan{\c{c}}arski.
\newblock A contextual-bandit algorithm for mobile context-aware recommender
  system.
\newblock In Tingwen Huang, Zhigang Zeng, Chuandong Li, and Chi~Sing Leung,
  editors, {\em Neural Information Processing}, pages 324--331, Berlin,
  Heidelberg, 2012. Springer Berlin Heidelberg.

\bibitem{6483433}
Y.~{Deshpande} and A.~{Montanari}.
\newblock Linear bandits in high dimension and recommendation systems.
\newblock In {\em 2012 50th Annual Allerton Conference on Communication,
  Control, and Computing (Allerton)}, pages 1750--1754, Oct 2012.

\bibitem{NIPS2017_6900}
Nir Levine, Koby Crammer, and Shie Mannor.
\newblock Rotting bandits.
\newblock In I.~Guyon, U.~V. Luxburg, S.~Bengio, H.~Wallach, R.~Fergus,
  S.~Vishwanathan, and R.~Garnett, editors, {\em Advances in Neural Information
  Processing Systems 30}, pages 3074--3083. Curran Associates, Inc., 2017.

\bibitem{doi:10.1002/asmb.874}
Steven~L. Scott.
\newblock A modern bayesian look at the multi-armed bandit.
\newblock {\em Applied Stochastic Models in Business and Industry},
  26(6):639--658, 2010.

\bibitem{6200864}
C.~{Tekin} and M.~{Liu}.
\newblock Online learning of rested and restless bandits.
\newblock {\em IEEE Transactions on Information Theory}, 58(8):5588--5611, Aug
  2012.

\end{thebibliography}

\end{document}